\documentclass[runningheads]{llncs}

\usepackage{verbatim}
\usepackage[utf8]{inputenc}
\usepackage[dvipsnames]{xcolor}
\usepackage{graphicx}
\usepackage{booktabs}
\usepackage{wrapfig}
\usepackage{hyperref}
\usepackage{float}
\usepackage{array}
\usepackage{setspace}
\usepackage{listings}
\usepackage[official]{eurosym}
\usepackage{graphicx}
\usepackage{amsmath}
\usepackage{amssymb}
\usepackage{floatflt}
\usepackage{makecell}
\usepackage{tikz}
\usetikzlibrary{shapes,arrows,positioning}
\hypersetup{bookmarks=true}

\newcolumntype{P}[1]{>{\centering\arraybackslash}p{#1}}



\title{Generating Reliable Process Event Streams and Time Series Data based on Neural Networks\protect\footnote{The final authenticated version is available online at https://doi.org/10.1007/978-3-030-79186-5\_6}}

\author{\small Tobias Herbert\inst{1}, Juergen Mangler\inst{2}, Stefanie Rinderle-Ma\inst{2}\\
\small
\small
\small tobias.herbert.uni@gmail.com, \\\small\{juergen.mangler, stefanie.rinderle-ma\}@tum.de}
\date{}

\institute{Faculty of Computer Science, University of Vienna, Austria
\and
Chair of Information Systems and Business Process Management, \\\small Department of Informatics, Technical University of \small Munich, Germany}

\begin{document}

\maketitle

\begin{abstract}

Domains such as manufacturing and medicine crave for continuous monitoring and analysis of their processes, especially in combination with time series as produced by sensors. Time series data can be exploited to, for example, explain and predict concept drifts during runtime. Generally, a certain data volume is required in order to produce meaningful analysis results. However, reliable data sets are often missing, for example, if event streams and times series data are collected separately, in case of a new process, or if it is too expensive to obtain a sufficient data volume. Additional challenges arise with preparing time series data from multiple event sources, variations in data collection frequency, and concept drift. This paper proposes the GENLOG approach to generate reliable event and time series data that follows the distribution of the underlying input data set. GENLOG employs data resampling and enables the user to select different parts of the log data to orchestrate the training of a recurrent neural network for stream generation. The generated data is sampled back to its original sample rate and is embedded into the originating log data file. Overall, GENLOG can boost small data sets and consequently the application of online process mining.
\end{abstract}

\section{Introduction}

\begin{figure}[h]
\includegraphics[width=\linewidth]{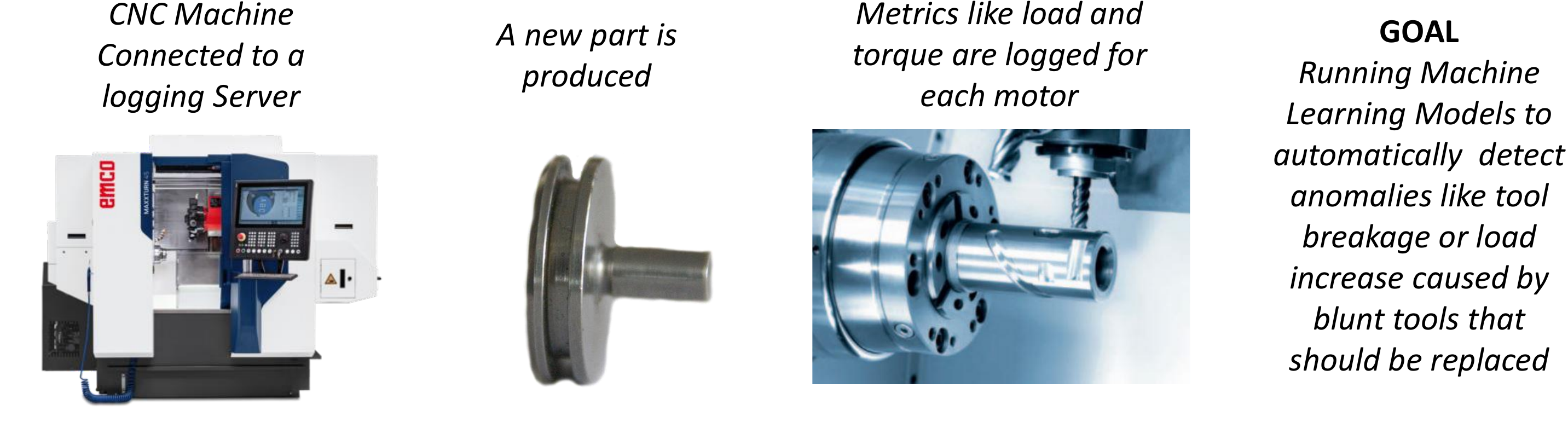}
\caption{Outages and machine deterioration inflict high costs in the manufacturing domain. The goal is to use machine learning models to detect tool deterioration and breakage to alert the machinist to avoid damages to the part, tools and the machine.}
\label{fig:domain_introduction}
\end{figure}

The continuous monitoring and analysis of their processes is crucial for many application domains such as manufacturing \cite{SMR2020} and medicine \cite{cs6535}. This holds particularly true if the processes are connected to physical devices such as machines or sensors that produce a plethora of data themselves, i.e., meta data that is actually not important for the genesis of work pieces and products, but is crucial to understand errors, optimize machines and processes, and to react to production problems more quickly. Often such machine and sensor data is collected as time series. Consider the manufacturing example depicted in Fig. \ref{fig:domain_introduction} where a CNC machine produces parts and the necessary tasks are logged into a \textsl{process event stream} during runtime. In addition, \textsl{metrics} such as load and torque are logged as time series for each motor. An analysis goal is to detect, explain, and predict anomalies in the process event stream (i.e., the process conducted by the CNC machine) based on the time series reflecting the parameters and using combined process mining and Machine Learning techniques\footnote{Previous work \cite{SMR2020} combines, e.g., online process mining with dynamic time warping.}.

As promising as the application of powerful analysis techniques sounds, the precondition is to be able to collect \textsl{reliable} data sets as input. By reliable we refer to data sets of a sufficient volume and quality. The quality aspect is tied with the volume aspect as, for example, removing data of low quality can take a toll on the volume. If, in turn, measures are taken to boost data sets artificially, reliability of the result should imply that the structure of the original data is maintained, i.e., the resulting data set is as realistic as possible. In real-world scenarios, the following reasons might hamper the creation of reliable data sets:

\begin{itemize}

    \item Slightly changing process models which often include manual repair steps, especially during early phases of production.

    \item Flaky time series data collection, as (a) machines utilize real-time computing capabilities and prioritize operational steps over data delivery steps, and (b) unstable networking and data aggregation, as individual sensors often produce more than hundred mebibytes per second.

\end{itemize}

For example the manufacturing domain, is in a tight spot. In theory it is expected to operate in a highly automated and connected manner, inline with widely advertised Industry 4.0 principles. In reality, it is still stuck with decade old machines, that are reliable, but black boxes when seen as sources of analyzable data, poor networking infrastructure, and lots of manual tweaking and repair steps. This results in gaps in most real-world manufacturing data sets.
And this does not only hold for the manufacturing domain. Problems hampering the creation of reliable data sets for analysis purposes, might occur in other domains (e.g., medicine or logistics) for exactly the same reasons.

The main goal of this work is to support the creation of reliable data sets, i.e., process event streams and time series data, in case of existing low volume data. In Machine Learning, synthetic data has been already generated according to a given distribution and purposes such as privacy\footnote{\url{https://research.aimultiple.com/synthetic-data-generation/}}. In this work, we propose GENLOG which employs Machine Learning techniques and is based on standard XES\footnote{XES stands for eXtensible Event Stream and is the de facto process log standard, \\\url{http://www.xes-standard.org/}.} log files. GENLOG aims at solving the following two problems:

\begin{itemize}

    \item \textbf{Problem 1:} The ex-post generation of solid data sets with slight variations regarding both - process events and time series data.

    \item \textbf{Problem 2:} Filling in gaps in data sets caused by variations in data collection frequency and other data collection errors or deficiencies.

\end{itemize}

Current approaches for generating log data (in a process setting) often require many parameters that have to be estimated/simulated based on probabilities or probability distributions. Moreover, existing approaches often do not focus on or even include time series data. In order to overcome these limitations, the GENLOG approach uses Neural Networks to generate event and sensor streams, the latter reflected as time series data, based on existing log data. The generated data is then embedded into the originating file type and structure. In particular, no parameters are required to be set by the user.

Nonetheless, GENLOG aims at reflecting the distribution of the input log data in the generated event and data streams.

The ideas behind GENLOG are underpinned by the following considerations: Machine Learning algorithms are in general an asset for analyzing, monitoring, and optimizing business processes \cite{DBLP:books/sp/Aalst16}. Whenever log data is available it is an opportunity to check for outliers and patterns \cite{OlinerAdam2012Aaci}, both for conformance and optimization purposes. Time series data can hold valuable insights about process execution, faults, failures, and changes in the process over time \cite{WeiWilliamW.S.2013TSA}. When a sufficient amount of data is available to train machine learning models, they can be used effectively in both online and offline scenarios such as near real time outlier detection as well as classification. However if there is only a small amount of data available it is not possible to train a model \cite{LecunYann2015Dl} that will generalize well to future data \cite{NarayanS2005Aaou}.

The implementation of GENLOG approach can be tried out at \url{https://cpee.org/genlog/}. The online tool allows to access all the data sets used in this paper, upload new ones, as well as access visual representations generated during the creation of new logs.

The remainder of this paper is structured as follows: Section \ref{sec:ldsg} describes how log data sets are typically structured, and how the GENLOG pipeline works on these data sets. Section \ref{sec:PI} outlines the prototypical implementation and Sect. \ref{sec:eval} details the evaluation in a real-world scenario. Section \ref{sec:related} presents related work. In Sect. \ref{sec:conclusion} the paper is concluded.

\section{GENLOG Pipeline for Creating Reliable Process Event Streams and Time Series Data}
\label{sec:ldsg}

Figure \ref{fig:domain_introduction} uses the manufacturing of new part as an example to illustrate a real-world scenario in which GENLOG can be added to improve the monitoring abilities of the existing system to detect machine malfunction and deterioration which both reduces the need for human intervention and can prevent machine outages if a tool head is replaced before it breaks or the quality of the final parts is out of specification.

Figure \ref{fig:current_approach_problem} picks up the scenario presented in Fig. \ref{fig:domain_introduction}. It describes how GENLOG can be used to deal with the
typical life-cycle of a part -- i.e. design, production, and iterative improvements. Improvements might be necessary due to machining or process inefficiencies, e.g. to detect tool breakage or deterioration. All of these cases initially lead to a situation where not enough data is available:

\begin{figure}[h]
\includegraphics[width=\linewidth]{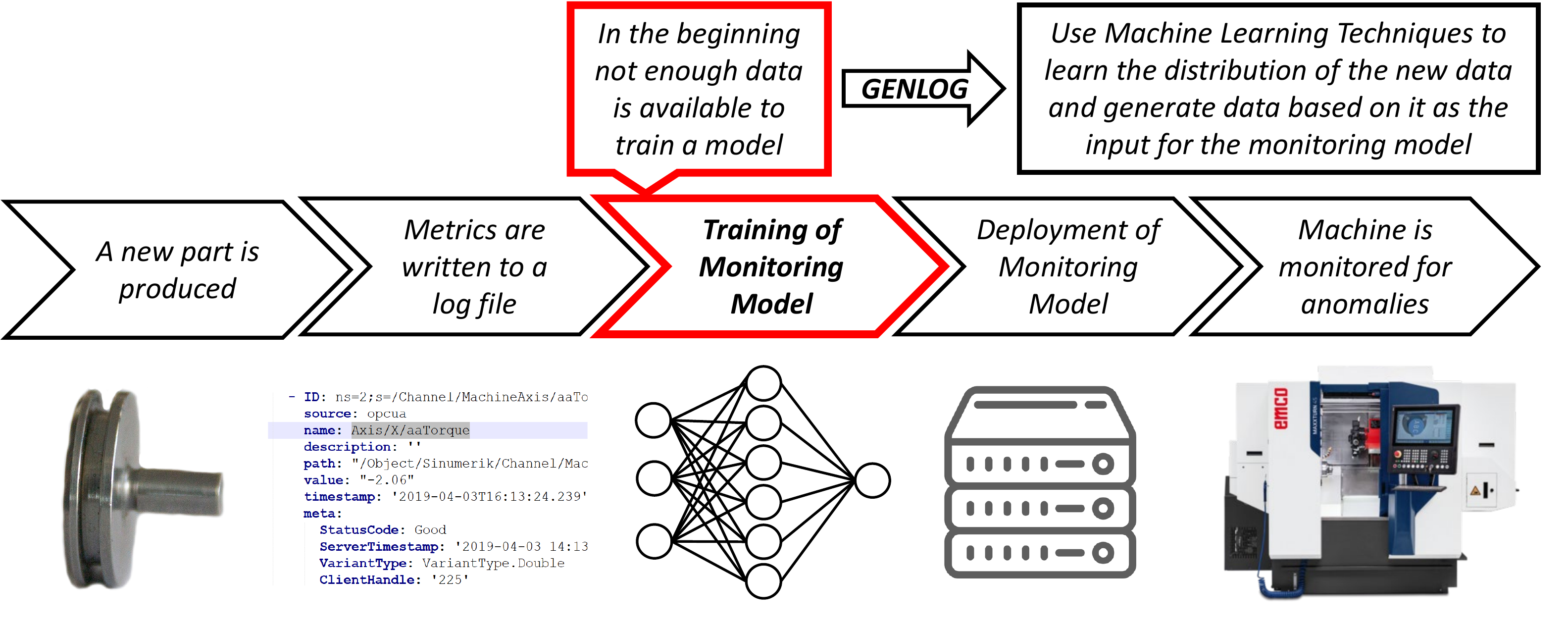}
\caption{Use GENLOG when a new part is produced and not enough training data is available as input for a Machine Learning Model}
\label{fig:current_approach_problem}
\end{figure}

\begin{enumerate}
\item Process mining and Machine Learning techniques are employed in existing analysis tool chains. It is desired to use them as soon as possible when a new process is introduced, which is sometimes not possible due to the small size of the data set.
In order to fill this ``data gap'', the GENLOG approach takes the existing data to generate new data that follows the same distribution and is of the same file type, structure, and frequency as the original data. This way the data generated by GENLOG can be immediately used in all existing analysis processes that take the log files as input data.

\item  Cleaning and pruning data for analysis purposes can also result in a data set that is too small to train a good model. Again the GENLOG approach can fill the gap by yielding a sufficient data set.

\item  Finally, the data might not be balanced, i.e., the amount of data instances in one class differs from other classes. This can cause a bias in the resulting model \cite{Fanny2018DLfI} which is also common in other domains, such as the medical sector (e.g., most test subjects are healthy). Instead of balancing the classes by throwing away data from larger classes, data can be generated for the smaller classes. This increases the ability of the model to correctly adapt to new data.
\end{enumerate}

GENLOG is added before the existing monitoring model is trained in order to provide a reliable data set, i.e., process event stream and connected time series. Thus from an early stage of production it is possible to train a model that can monitor the system and it will improve over time with more real data available for both direct training and as input to GENLOG.

\begin{figure}[h]
\includegraphics[width=\linewidth]{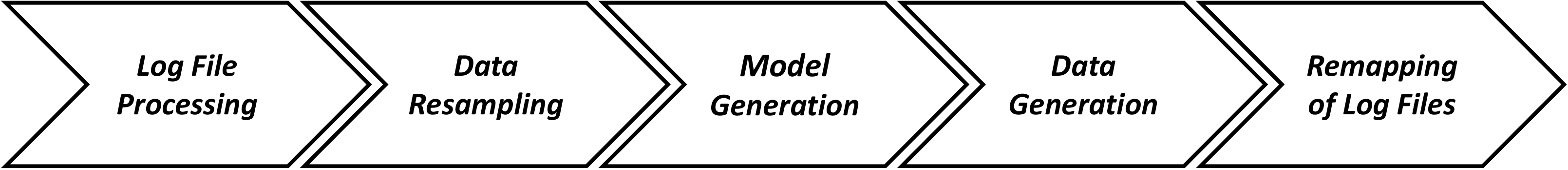}
\caption{GENLOG steps taking a log file as input, training a Machine Learning model to generate data and creating an output log file including the generated data}
\label{fig:genlog_approach}
\end{figure}

Fig.~\ref{fig:genlog_approach} shows the pipeline of the GENLOG approach. In this section, we start by sketching how the pipeline is designed in general, followed by details on the employed Machine Learning concepts. In Sect. \ref{sec:PI}, the description of the GENLOG pipeline is fleshed out in the context of its prototypical implementation.

The GENLOG pipeline starts by processing the input log files. The result of this step is a collection of time series data, one for each metric for each log file. Examples for metrics are sensor data for temperature, acceleration or power related measurements. Next the data is resampled to a common sample rate for further processing. The data is then ready to be used as input for the training of the Neural Networks. At this stage, no human intervention was necessary and the only parameter was the list of metrics that should be extracted from the logs which can also be omitted if all metrics should be used. Now the user is presented with a dashboard that shows the metrics in terms of its grouping either by batches or temporal which are used to choose the data that will be the base for the data generation. The data is then generated using the prediction method of the trained models and using existing data as input. At this stage the data is still a time series and the final and crucial step is to map it back into its original file format. Thus it is embedded back into the original log file which results in a new log file with generated data that can be used by any process that uses log files as inputs and it could even be used in GENLOG to create new models which has been done and can be evaluated in the online tool.

\begin{figure}[h]
 \begin{center}
\includegraphics[width=0.8\linewidth]{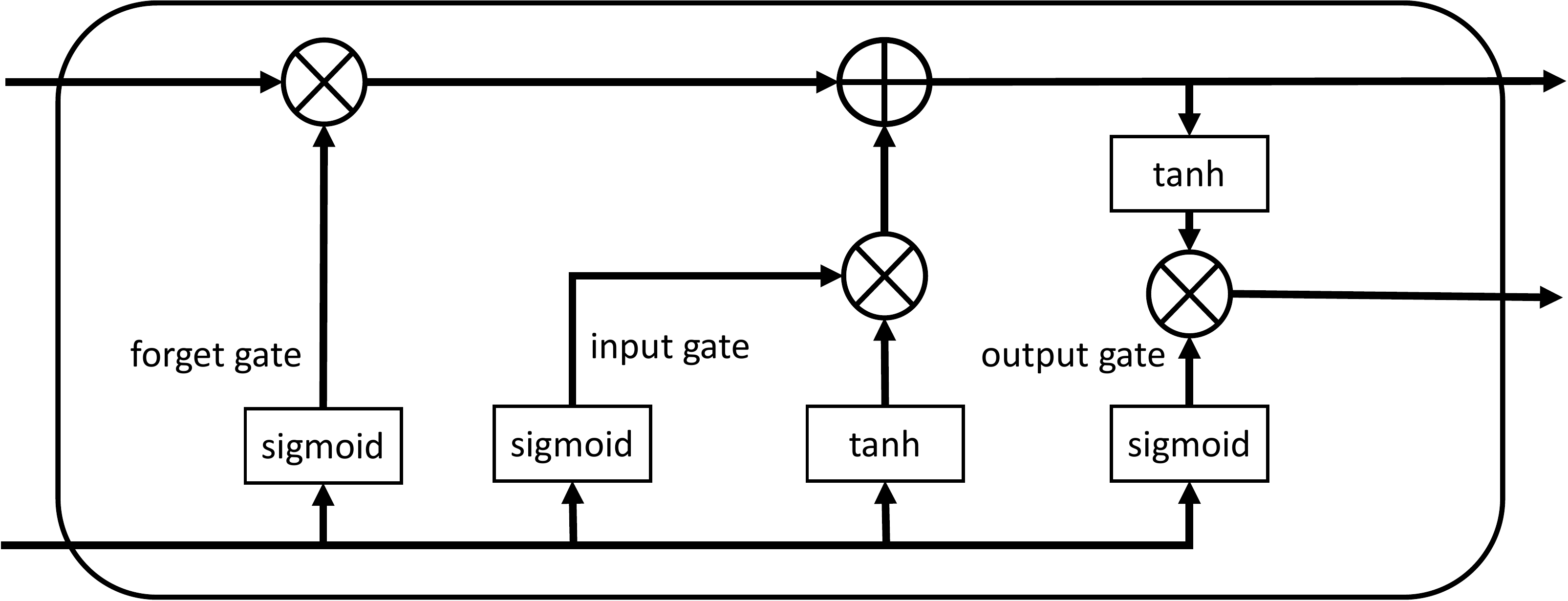}
\end{center}
\caption{Conceptual Illustration of an LSTM}
\label{fig:lstm_concept}
\end{figure}

Standard Neural Networks can be used in many applications. They do however not have a concept of time, but rather see their input as just a collection of inputs without a temporal relationship. Recurrent Neural Networks (RNNs) \cite{MandicDaniloP2001RNNf} use a hidden state vector to create a temporal context which influences the output of each node together with the weights of the node. This makes it a powerful concept for data generation and time series prediction, but they are still not able to keep important information over a longer period of time. During back propagation, recurrent neural networks suffer from the vanishing gradient problem \cite{PascanuRazvan2012Otdo} which causes the updates of the weights to become smaller the further the back propagation progresses which hinders learning. A Long Short-Term Memory (LSTM) \cite{10.1162/neco.1997.9.8.1735} addresses this issue by keeping a global state and using gates to determine what information should be forgotten and what should be kept. It uses the hyperbolic tangent function to regularize the data and the sigmoid function to create forget, input and output gates that control which information are worth keeping and what should be discarded as shown in Fig.~\ref{fig:lstm_concept}\cite{lstm_concept}. Depending on the use case, it is possible to train many LSTM models even down to the individual instances. It is also possible to aggregate the data first and then train a model on, e.g., the median and standard deviation to reduce the effect of noise in the data. Besides the trained models, additional variance can be achieved by using the existing data as input for the models predict method such that the generated data is a combination of model and input data. By using domain knowledge it is possible to control the distribution of the data generation in the user interface by omitting invalid or problematic data. In the GENLOG approach, models are trained at fine granularity and the user selects the data that should be used to generate new log data. It can also be implemented as a continuous training by using new data to train new models and thus over time both replacing artificial data with real data and improving the quality of the generated data by having access to more real data for training. This approach allows to train machine learning models from a much earlier stage where only a small amount of data is available and over time increase the quality of the prediction. This is especially relevant to domains which deal with new scenarios regularly like the manufacturing domain where new parts are produced with no initial data available or the process model changes.

\section{Prototypical Implementation}
\label{sec:PI}
The GENLOG pipeline depicted in Fig. \ref{fig:genlog_approach} and all the processing steps, model training, and data generation are implemented using python. The frontend makes heavy use of d3.js which is used to dynamically create the user interface based on the batches and metrics and is used to plot the data for validation and visual data analysis. The LSTM is implemented with the Tensorflow \cite{ZacconeGiancarlo2017DLwT} backend which uses CUDA \cite{ShiShaohuai2016BSDL} for GPU accelerated training. The repository for the tool can be found here\footnote{https://github.com/tohe91/GENLOG} and the tool can be used for evaluation here\footnote{https://cpee.org/genlog/}. The core idea is to automate as much as possible and at the same time provide users with a high degree of influence over what data is used as the base for data generation, i.e., by removing or adding metrics and changing the mapping. This flexibility enables users to react especially to concept drift \cite{DBLP:journals/tnn/BoseAZP14}, i.e., changes in the log data that reflect process change and evolution.

\subsection{Log File Processing}
This step resembles an interface between the existing log file generation processes \cite{DBLP:conf/icsoc/StertzRHM16,ManglerJuergen2010CPEE} and the test set data generation pipeline. The time series data for different metrics is extracted from the log files and stored in csv files as a timestamp and value pair. In a heterogeneous environment there can simply be multiple interfaces that take different input formats like XES and MXML \cite{from2005meta} and output the time series csv files.

\subsection{Data Resampling}
All the data is now in the same format, but the frequencies differ both between the different measured metrics and within one file. The latter often arises because log data is written by the same device that is executing the process which can take resources away from log creation and towards the execution of the main process as well as data being transmitted through a network. The scale of the values matters in multivariate data as the metrics together form the features for model training. In a univariate model, feature scaling is not necessary as only one metric is used as the feature. To train a multivariate model it is advised to normalize the data because some loss metrics can introduce a bias otherwise \cite{jayalakshmi2011statistical} or converge more slowly as it is the case with gradient descent.

\subsection{Model Generation}
Preparing the data for training of the LSTM is done by creating a feature and a target vector. The target vector is created by shifting the values of the feature vector such that the target value is always the next value relative to the feature vector which is what the model is supposed to predict. The model is created by defining the size of the LSTM layer and its activation function. The activation function used is the Rectified Linear Unit (ReLU) \cite{HaraKazuyuki2015Aofo} which provides better gradient propagation \cite{pmlr-v15-glorot11a} together with Adam optimization \cite{kingma2014adam}. The size of the output layer is the number of features used and is one for a univariate time series. The loss function used is the Mean Squared Error \cite{CharkhiAli2015Mmse}. A callback for early stopping \cite{BlanchardGilles2018Oafe} is added to increase performance by monitoring the loss function and stopping the training before max iterations are reached if the loss function did not change significantly in the last number of iterations.

\subsection{Data Generation}
To generate new data, both a trained model and an original data instance are chosen uniformly at random from a set that can be defined by the user through the user interface and the predict method of the model is called with a original data instance as the input. What makes this approach so powerful are the selective options that the user has. Often data can be grouped temporally by changes in the process and thus create data collections that correspond to a specific state of the process. By grouping the input data and the resulting trained models by the state of the process and other groupings like batch production or medical trials, the user can choose what should be the base for the newly generated data. This can be achieved by choosing the state of the process for each feature individually. In case there is no natural grouping apparent in the data, the user can chose a time window for each feature to represent the data basis for the data generation. This gives the user fine grained control over the generated data which can be beneficial if it is known that a specific feature was invalid or undesired for a given state of the process. Another approach the user can take is to specify if the generated data should be similar to the most recent data or if it should cover a broader time span. This can eliminate the need for manual pruning of data that is often required when using data analysis algorithms on raw log data. It can also be useful if manual pruning results in a data set that is too small for use in data analysis algorithms to produce meaningful results. By taking the pruned data and generating more data based on its distribution we create a more reliable data set.

\subsection{Remapping to Log File Format}
In order to go from validating the generated data to using it in existing workflows it is important to map the data back to its original log format. As we have extracted and resampled the data in the beginning of the pipeline we now have to reverse these steps by sampling it back and by embedding it into the original log file. To embed each metric, first the time series for that metric has be extracted and becomes the template for the generated data. The generated data is then resampled to match the template which can lastly be embedded back into the original log file. This makes it possible to maintain the integrity of the log file and selectively adding generated data to it. A detailed explanation will be given in the next section where the GENLOG approach is applied to the production of a metal part from a CNC turning machine on the shop floor.

\section{Evaluation in a Real World Scenario}
\label{sec:eval}
\begin{figure}[h]
\includegraphics[width=\linewidth]{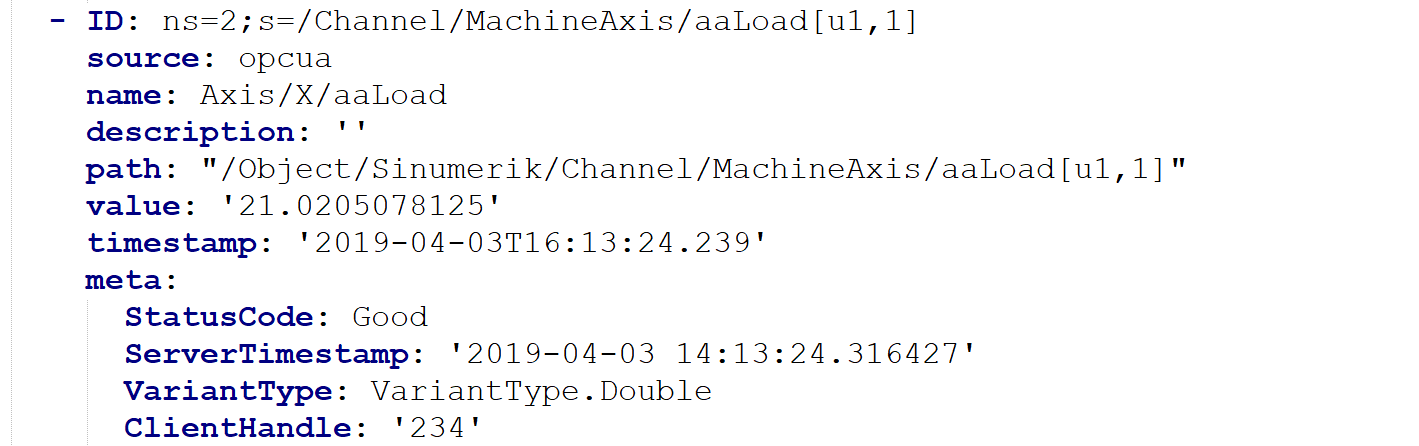}
\caption{YAML Log file produced in the GV12 Production}
\label{fig:yaml}
\end{figure}

The GENLOG pipeline (cf. Fig. \ref{fig:genlog_approach}) is applied in a production environment based on log data from the Competence Center for Digital Production (CDP)\footnote{http://acdp.at}. An EMCO ``MaxxTurn 45 SMY'' machines a part in the GV12 production. GV12 is a \textsl{``part of a valve from a gas motor. The complexity of this product is caused by the small tolerances, which necessitates the complex overall scenario''} \cite{DBLP:conf/bpm/ManglerPRE19}. The production process is executed and logged by the Cloud Process Execution Engine (CPEE) \cite{ManglerJuergen2010CPEE,DBLP:conf/icsoc/StertzRHM16}. The CPEE also logs metrics of the machine like the torque and load of the motors for each axis as well as spindle speed, spindle load, and other metrics from the turning process together with the process execution steps and meta data \cite{DBLP:conf/wecwis/EhrendorferFMR19}. The produced log file originates in XES format and is then stored as a YAML file for further usage (cf. Fig. \ref{fig:yaml}).

\subsection{Log File Processing}
The created log files (as shown in Fig. \ref{fig:yaml}) include all the metrics and information about the state of the machine, the state of the process, and meta data from the servers and production environment. The only requirement for the GENLOG pipeline to run is to provide the log files and their internal structure if it differs from the default and enter the metrics that should be extracted. If no metrics are specified all of the available ones are selected by default, i.e., the user does not have set any parameters. The pipeline then handles the extraction, resampling, and model generation without supervision or further input.

\subsection{Data Resampling}
The data is extracted and stored as time series data for each metric. The GENLOG tool enables the user to use visual analysis to find outliers, view different aggregations of the data, and analyze how the metrics change over time and through changes in production by plotting the data. The production of the parts is organized in batches which creates a natural grouping that is used to organize the log data. In the background the data is resampled to a common sample rate which is at least as high as the highest rate between all the metrics to make sure no data is lost in the resampling process.

\subsection{Model Generation}

\begin{figure*}[h]
\includegraphics[width=\linewidth]{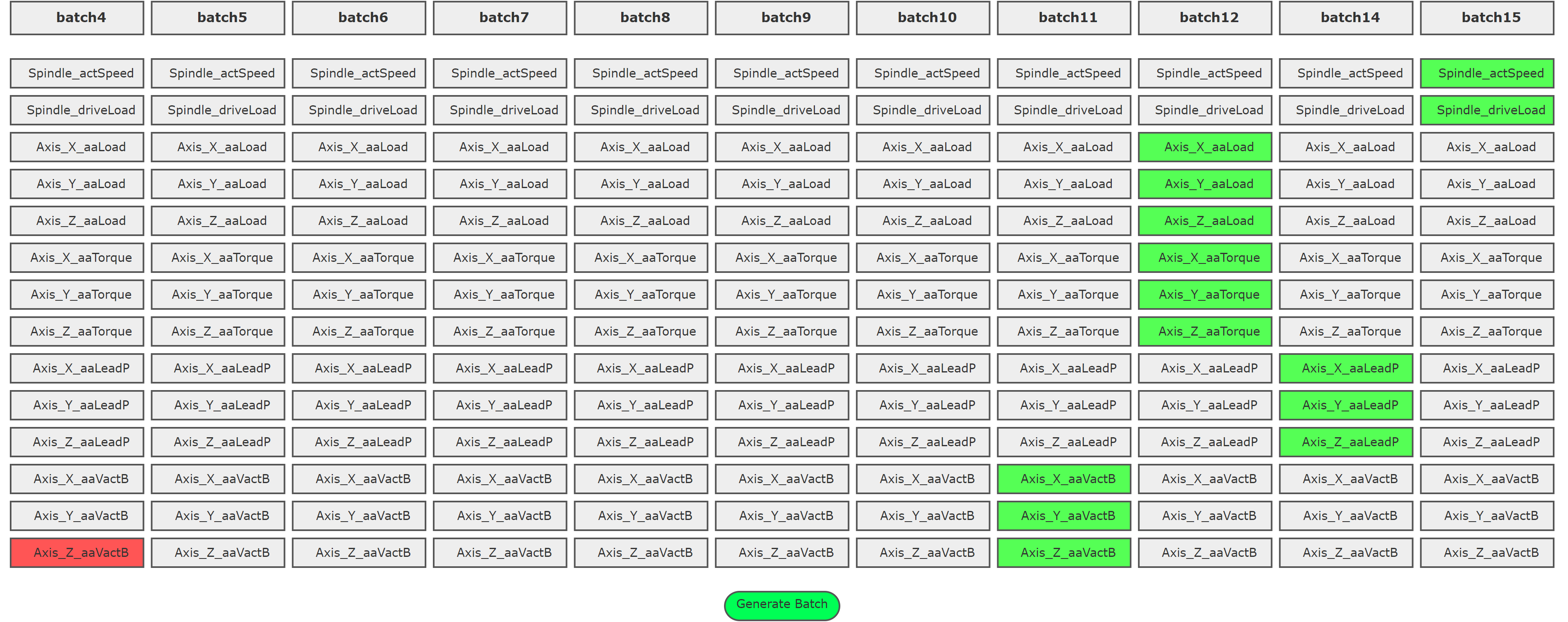}
\caption{User Interface for data generation, \textbf{red}: model not trained yet - \textbf{gray}: model ready -  \textbf{green}: model active for data generation}
\label{fig:ui}
\end{figure*}

The models are generated automatically as part of the pipeline and a fine grain model generation is used to preserve the characteristics of the individual production runs and no aggregation or generalization is done. Using an LSTM in this context is powerful as the different metrics have very different characteristics with some following a low polynomial function while others follow a  high polynomial function with a high number of oscillations. This can be covered well by the LSTM because it predicts one data point at a time by using both recent information like an RNN and also older information which gives it its long term memory characteristics. Adding the early stopping functionality increases the total run time of the model creation significantly as some metrics result in no changes in the loss function earlier than others. For reference the creation of over 5800 LSTM models took just 5 hours on a standard laptop (i7-7700HQ CPU, 16GB RAM, NVIDIA GeForce GTX 1060).

\subsection{Data Generation}
At this point the user is presented with a dashboard (see Fig.~\ref{fig:ui}) showing all the metrics for the individual batches. In the generic case this is where the temporal window could be chosen if there was no grouping in batches or process changes made beforehand. The user interface shows for which metric-batch combination the models have already been trained and are ready to use. The user can then pick which batches should be the data basis for each of the metrics. In case only the most recent data should be used as the basis the user can simply select the latest batch which activates all metrics for this batch and the generated data will represent this batch in all metrics. In case of a new process that only has very few data produced at this point, the selection can be based on the entire available data. In the scenario that multiple batches are chosen, the individual models and data instance, which represents one produced part, are chosen uniformly at random and are combined at random. As an example we chose batch14 and batch15 for a metric, there can be four possible combinations, model14-data14, model-14-data15, model15-data14, model15-data15, which produces a high degree of variance even if the original data set size was small because we can combine models and data input together.

\subsection{Remapping to Log File Format}
The previous step is sufficient to validate how well the newly generated data represents the same distribution as the original data. In order to become useful for data analysis, the data needs to be embedded into the original file format. In the remapping stage the generated data is embedded into the original log file. The embedding is done by first calculating the duration of the process from the original log file in milliseconds. Second the number of logged values for each metric are summed up, thus we have the duration of the process and the number of logged values for each metric which allows us to calculate a sample rate for each metric. Third the generated data of each metric is resampled with the calculated sample rate and therefore the generated data has the same amount of entries as the original log file for each metric. Fourth both value and timestamp for each metric are embedded into the log file. The timestamp was created by taking the first timestamp of the log file and then using the sample rate to increase each timestamp accordingly. As a result the generated log file has the same structure, length, variance and meta data as the originating log files.

\begin{figure}[h]
\includegraphics[width=\linewidth]{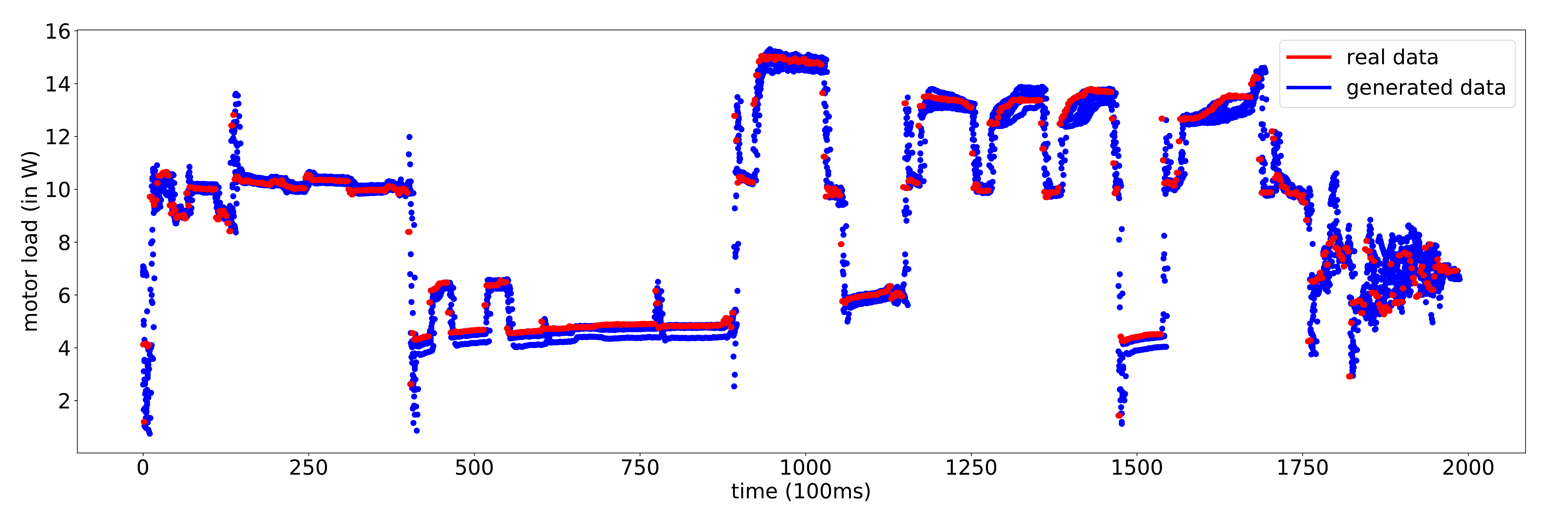}
\caption{The real data is shown in red while blue shows the distribution of 100 generated data samples for the load of the Y Axis Motor in the turning machine}
\label{fig:result2}
\end{figure}

\subsection{Discussion of the results}
\begin{figure}[h]
\includegraphics[width=\linewidth]{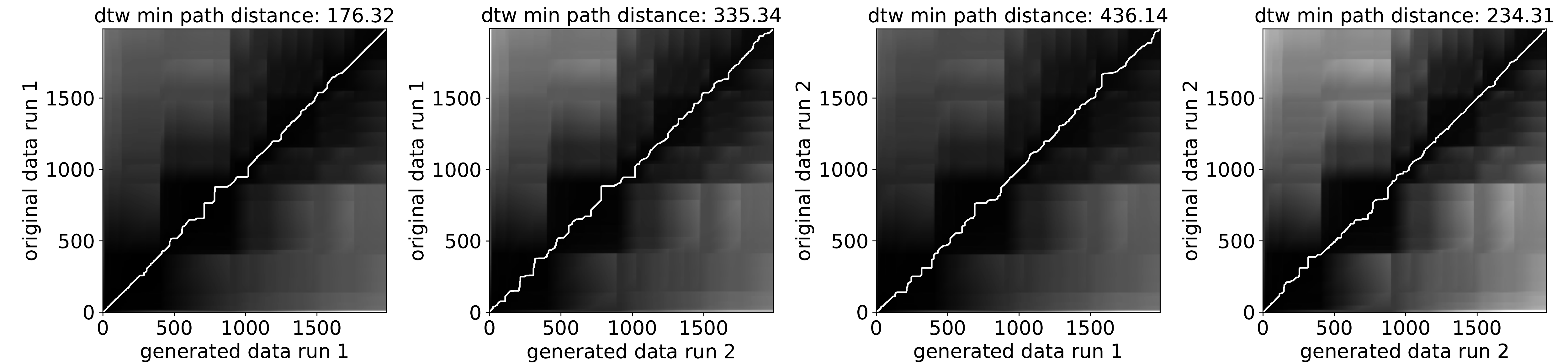}
\caption{Minimum path for dynamic time warping crossing two models with their originating time series data}
\label{fig:dtw}
\end{figure}

The results are encouraging as Fig.~\ref{fig:result2} shows the plots for generated data representing the original data quite well. Our technique is very suitable for processes that experience concept drift, meaning the process is subject to change over time, as the LSTM is conditioned to put emphasis on more recent data whilst considering the full data set. As a result the models continuously evolve with the process without the need to make a hard cut and starting over with an again very small data set that only represents the altered process. It might be necessary to reset the data set in case the changes to the process are too severe, like changing the order of machining steps performed on the part, as this might cause unpredictable and undesirable artifacts in the generated data. Another limitation of this approach becomes apparent when a very small data set is used that does not include any significant variance. If the available data is very biased and not representing the real distribution well, the amount of variance GENLOG can introduce is also limited. Therefore we plan to introduce a list of statistical measures like variance and standard deviation and using dynamic time warping on the original data set and the resulting boosted data set to understand better how much variance can be introduced on different data sets.
Fig.~\ref{fig:dtw} shows the minimum path for dynamic time warping between the generated data from two models and the two originating time series. The results show that using its own time series as input, the added variance on the output is small but exists because the models are not overfitted and therefore have not learned the data perfectly and when using the time series from another run, the added variance increases. This shows how cross using all models with all time series allows to boost a very small data set with realistic data.

\section{Related Work}
\label{sec:related}

The process mining manifesto states that the provision of high quality event data is crucial (Guiding Principle 1 and Challenge 1) and that process mining should be combined with other analysis techniques (Challenge 9) \cite{DBLP:conf/bpm/AalstAM11}. GENLOG contributes to both claims.
The generation of process log data can be approached from two directions, i.e., through simulation based on process models (e.g., using Colored Petri Nets \cite{DBLP:journals/eis/MaggiW17} with CPN tools\footnote{\url{http://cpntools.org/}}) or through executing process models in ``test mode'', e.g., with mocked services \cite{DBLP:conf/icsoc/StertzRHM16}. These approaches, however, rely on process models with estimated values, e.g., probabilities at alternative branchings, and do not generate sensor data. An approach for integrating event streams from heterogeneous data sources is presented in \cite{DBLP:conf/icpm/KoenigMR19}, but no event streams are generated. Approaches such as \cite{SMR2020} analyze and exploit event and sensor streams, but do not generate them either. Analyzing the capability of GENLOG for predicting the next activity in a process execution in comparison to existing approaches such as \cite{EvermannJoerg2017XT-P} using deep learning is part of future work.

LSTMs are not limited to time series data and can also be used for sequence labeling \cite{MaXuezhe2016ESLv} and text classification \cite{ZhouChunting2015ACNN} which can be used on the process and task level for conformance checking and anomaly detection. Gated Recurrent Units which are related to LSTMs and are also based on RNNs can be used to predict bottlenecks \cite{FangWeiguang2020APGR} in a production setting. So RNN based models might also be useful to address some of the issues in process mining that can be prone to both overfitting and underfitting \cite{AalstW.2010Pmat} by creating models that use existing data instead of creating rules and using parameters.

 In future research we intend to compare the results of the LSTM with different Generative Adversarial Networks (GAN) \cite{goodfellow2014generative} variants within GENLOG to understand both theoretical and practical implications for both approaches for the generation of time series data. Grid search cross validation will be employed on all models to make it easier to adjust to new data sets without user intervention as a set of training parameters will be exhaustively checked and the best performing combination will be chosen as part of the pipeline.

\section{Conclusion}
\label{sec:conclusion}

A prototypical implementation of GENLOG is used offline to generate new event and time series data to be used as input for analysis algorithms. We applied GENLOG to a real-world data set from the manufacturing domain. The results showed that the models are not suffering from overfitting or underfitting as they introduce realistic variation both for their originating time series input as well as for the input of other runs.

Overall, the results show that it is feasible to use a pipeline such as GENLOG which does not rely extensively on pre-configured parameter sets, but instead solely utilizes historic log data, in order to generate new time series log data.

\noindent\textbf{Acknowledgements:} This work has been partly funded by the Austrian Research Promotion Agency (FFG) via the "Austrian Competence Center for Digital Production" (CDP) under the contract number 854187. This work has been supported by the Pilot Factory Industry 4.0, Seestadtstrasse 27, Vienna, Austria.

\bibliographystyle{abbrv}
\bibliography{out}
\end{document}